\documentclass[sigconf]{acmart}

\def\BibTeX{{\rm B\kern-.05em{\sc i\kern-.025em b}\kern-.08emT\kern-.1667em\lower.7ex\hbox{E}\kern-.125emX}}
    
\copyrightyear{2019}
\acmYear{2019}
\setcopyright{acmlicensed}
\acmConference[KDD '19]{August 04--08, 2019}{Workshop on AI for fashion}{Anchorage, Alaska-USA}
\acmBooktitle{KDD '19, Workshop on AI for fashion}
\acmPrice{15.00}

\begin{document}

%
\title{$c^+$GAN: Complementary Fashion Item Recommendation}

%




\author{Sudhir Kumar}
\affiliation{%
 \institution{Microsoft AI \& R}
 \city{Hyderabad}
 \state{Telangana}
 \country{India}}
 \email{sudkum@microsoft.com}
 
 \author{Mithun Das Gupta}
\affiliation{%
 \institution{Microsoft AI \& R}
 \city{Hyderabad}
 \state{Telangana}
 \country{India}}
 \email{migupta@microsoft.com}
 




%
\renewcommand{\shortauthors}{Kumar and Das Gupta}

%
\begin{abstract}
We present a conditional generative adversarial model to draw realistic samples from paired fashion clothing distribution and provide real samples to pair with arbitrary fashion units. More concretely, given an image of a shirt, obtained from a fashion magazine, a brochure or even any random click on ones phone, we draw realistic samples from a parameterized conditional distribution learned as a conditional generative adversarial network ($c^+$GAN) to generate the possible pants which can go with the shirt. We start with a classical cGAN model as proposed by Mirza and Osindero~\cite{MirzaO14} and modify both the generator and discriminator to work on captured-in-the-wild data with no human alignment. We gather a dataset from web crawled data, systematically develop a method which counters the problems inherent to such data, and finally present plausible results based on our technique. We propose simple ideas to evaluate how these techniques can conquer the cognitive gap that exists when arbitrary clothing articles need to be paired with another relevant article, based on similarity of search results.
\end{abstract}

%
%


%

\begin{teaserfigure}
\centering
\includegraphics[width=16cm,height=7.5cm]{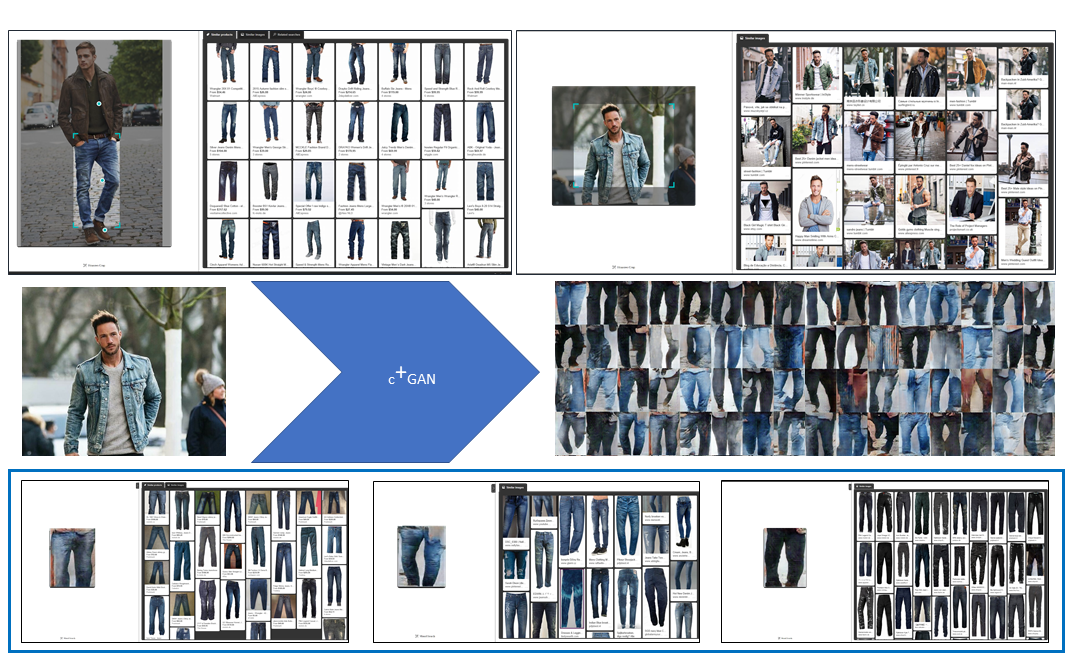}
\caption{For full body fashion images, current day image search can find matches for the individual clothing items. This is not the case when the user is looking for complementary information, like the choice of pants for his jacket, as shown in the top right panel. The image search engine goes to its default mode and finds more tops. The middle panel shows the GAN pipeline developed in this paper to learn a conditional mapping from the top images to the bottom images. The bottom panel shows the retrieval of matching articles based on the generated bottoms.}
\label{Fig:Prob1emDef}
\end{teaserfigure}

%
\maketitle

\section{Introduction}
The fashion domain is fast becoming a confluence of research ideas for computer vision. New problems involving key concepts of computer vision are emerging in tandem with the fashion industry’s rapid evolution towards creating an online, social, and personalized fulfillment model. Recent development in this field, namely style models~\cite{HsiaoG17,kiapourECCV14hipster,Lee2017Style2VecRL,MatzenBS17,SimoSerra2016FashionSI}, 
trend forecasting~\cite{AlHalah2017FashionFF}, interactive search~\cite{Kovashka2012WhittleSearchIS,Zhao2017MemoryAugmentedAM}, and recommendation~\cite{HuYD15,Lin2018ExplainableFR,Liu2012,Veit2015} 
all require visual understanding of fashion data with rich detail and subtle context. Research in this area has the promise to directly influence what people buy, how they shop, and how the fashion industry changes its paradigms and grows along with the research directions.

Fashion recommendation plays an increasingly important role in the online retail market. The purpose of fashion recommendation is to promote people's interest and participation in online shopping by recommending fashionable outfits that they may be interested in. For now, information retrieval (IR) based recommendation dominates the fashion market, where other similar users buying similar things together is considered the strongest signal to recommend an item. Early studies on fashion recommendation are based on small but expert-annotated datasets~\cite{Iwata2011FashionCR,Liu2012}, which cannot be directly translated to the development of complex models that need large amounts of training data (e.g., deep learning (DL) based models). One prominent push towards the creation of these large, web scale data is the 
proliferation of fashion-oriented online communities, e.g., Polyvore\footnote{http://www.polyvore.com/} and Chictopia\footnote{http://www.chictopia.com/}, where people can not only consume recent fashion trends but also share and comment on outfit and accessory compositions. 
In addition to a large number of outfit compositions, such crowd-sourced data also contains valuable information (e.g., user comments) for building more accurate and intelligent recommender systems.

In this paper, we address the task of complementary outfit recommendation based on a conditional representation of the clothing items. Given a top (i.e., upper garment), we need to recommend a selection of bottoms (e.g., trousers). We move away from the more commonly attempted IR based solution and actually try to draw a sample from the conditional distribution $p(x|y)$ of the lower garment $x$ given the upper garment $y$. These samples are learned to be realistic looking which can then be used to fetch purchasable items from any image search / e-commerce platform. We perform the conditional distribution estimation and sampling from a conditional generative adversarial network denoted as ($c^+$GAN) with realistic generation constraints. The proposed $c^+$GAN is built by incorporating multiple enhancements over the original cGAN method by Mirza et al.~\cite{MirzaO14}. 
The $c^+$GAN generates new samples from the learned conditional distribution, while maintaining strong correspondence with the underlying patterns, thereby making it somewhat easier to explain why the top and the bottom match. 


\section{Related Work}
No previous work has proposed a model similar to $c^+$GAN to generate fashion items starting with minimum manual intervention at scale. We briefly survey related work on fashion recommendation and on explainable recommendation, respectively and the move onto defining the integral parts of the work proposed in this paper.

\subsection{Fashion recommendation}
Given a photograph of a fashion item (e.g., tops), a fashion recommender system attempts to recommend a photograph of other fashion items (e.g., bottoms). This simple definition does not take into account whether the photograph contains only the apparel images or it contains a model wearing that apparel. The second scenario is more difficult than the first one. We tackle the second situation, but many of the large scale systems tackle the first scenario. 
Iwata et al.~\cite{Iwata2011FashionCR} propose a probabilistic topic model to recommend tops for bottoms by learning information about coordinates from visual features in each fashion item. Liu et al.~\cite{Liu2012} study both outfit and item recommendation problems. They propose a latent Support Vector Machine (lSVM) based model for occasion-oriented fashion recommendation, i.e., given a user-input occasion, suggesting the most suitable clothing, or recommending items to pair with the reference clothing. Jagadeesh et al.~\cite{Jagadeesh2014} propose two classes of fashion recommenders, namely deterministic and stochastic, while they mainly focus on color modeling for fashion recommendation.

Several recent publications have moved away from smaller manually annotated datasets and resorted to other sources, where rich data can be harvested automatically. One such effort in the personalized whole outfit recommendation space is by Hu et al.~\cite{HuYD15}, who propose a functional tensor factorization method to model interactions between users and fashion items over a dataset collected from Polyvore. McAuley et al.~\cite{McAuleyTSH15} employ a general framework to model human visual preference for a pair of objects of the Amazon co-purchase dataset. Their primary idea is to extract visual features by using deep convolutional neural nets (CNNs) and then introduce a similarity metric to measure visual relationships. Similarly, He and McAuley~\cite{He2016VBPRVB} introduce a scalable matrix factorization approach that incorporates visual signals into predictors of people\textquotesingle s opinions. To take contextual information (such as titles and categories) into consideration, Li et al.~\cite{Li2017MiningFO} classify a given outfit as popular or non-popular through a multi-modal and multi-instance deep learning system. To aggregate multi-modal data of fashion items and contextual information, Song et al.~\cite{Song2017NeuroStylistNC} first employ an auto-encoder to encode the latent compatibilities existing between the two spaces, and then employ Bayesian personalized ranking to exploit pairwise preferences between tops and bottoms. Han et al.~\cite{Han2017LearningFC} propose to jointly learn a visual-semantic embedding and the compatibility relationships among fashion items in an end-to-end manner. They train a bidirectional LSTM model to sequentially predict the next item conditioned on previous ones. 
Lin et al.~\cite{Lin2018ExplainableFR} propose a neural fashion recommender (NFR) model that simultaneously provides fashion recommendations and generates abstractive comments. 

The problem we want to solve in this paper is demonstrated in Fig.~\ref{Fig:Prob1emDef}. We posit that it is easy to find matching pieces from a database when the user has some idea what she is looking for. Many a times knowing what one is looking for is difficult. A conditional GAN can be trained to fill up this missing piece by generating possible clothing choices which can be subsequently used for searching over an e-commerce or fashion search platform. The proposed 
model learns to sample from the conditional distribution of $p(top | bottom)$ or the reverse one is used to generate plausible samples from this distribution. These samples are then used as search queries to match the users search intentions.

\subsection{Contributions}
In this work we present an end to end system to generate fashion recommendations for bottom apparels given a particular top clothing item. More specifically:
\begin{itemize}
    \item We propose a novel perceptual loss function using a simple discrete cosine transform (DCT) which is extremely fast and captures most of the high resolution details needed for image reconstruction,
    \item We propose a novel discriminator cost function called randomized label flipping, which is similar in ideology to the lens network proposed in~\cite{temperedAdvNets}, but with a single parameter which requires no additional training data by itself,
    \item Propose a scalable model for fashion recommendation with data gathered from the web. We use existing web based tools for data cleaning, which makes it extremely simple for other researchers to reproduce our work.
\end{itemize}

\section{$c^+$GAN Model}
Conditional GANs were introduced by Mirza and Osindero~\cite{MirzaO14} where in they fed the data $\mathbf{y}$ on which they want to condition the generation process to both the generator and the discriminator. Basic GAN as proposed by Goodfellow et al.~\cite{GAN_NIPS2014} can be  extended to a conditional model if both the generator and discriminator are conditioned on some extra information $\mathbf{y}$. The extra information encoded by $\mathbf{y}$ could be any kind of auxiliary information, such as class labels or data from other modalities. We can perform the conditioning by feeding $\mathbf{y}$ into the both the discriminator and generator as additional input layer. In the generator the prior input noise $p_z(\mathbf{z})$, and $\mathbf{y}$ are combined in joint hidden representation, and the adversarial training framework allows for considerable flexibility in how this hidden representation is composed. In the discriminator $\mathbf{x}$ and $\mathbf{y}$ are presented as inputs and to a discriminative function.
The objective function of a two-player minimax game is shown in Eq.~\ref{EQ:cGANCost}.
\begin{eqnarray}\label{EQ:cGANCost}
    \underset{G}{\min} ~ \underset{D}{\max} ~ V(D,G)  &=& E_{\mathbf{x} \sim p_{d}(\mathbf{x})}[\log D(\mathbf{x}|\mathbf{y})]                  \\
    \nonumber
    &+& E_{\mathbf{z}\sim p_z(\mathbf{z})}[\log(1 - D(G(\mathbf{z}|\mathbf{y})))]
\end{eqnarray}

In our work the fashion images, which are the images where a human model is visible with both top and bottom apparels, are segmented to represent the two inputs to the system $\mathbf{x}$ and $\mathbf{y}$. Since $\mathbf{x}$ represents the actual image to be generated it is encoded as an RGB image of size $224 \times 224$. The condition input $\mathbf{y}$ is encoded by passing the image through a deep network, Resnet-50 in our case, to generate a 1000 dimensional vector. The noise dimension is a hyperparameter in our model. Our goal for the generator is to train a function G that estimates, for a given conditional image feature $\mathbf{y}$ as input, its corresponding fashion image counterpart $\mathbf{x}$ as output. To achieve this, we train a generator network as a feed-forward CNN, $G_{\theta_G}$ parameterized by $\theta_G$. Here $\theta_G = \{W_{1:L}; b_{1:L} \}$ denotes the weights and biases of a $L$-layer deep network and is obtained by optimizing a conditional fashion image generation specific loss function denoted as $l_{cf}$. For training pairs $\mathbf{x}_i, \mathbf{y}_i$ for $i \in [1,\ldots, N]$, we solve
\begin{eqnarray}\label{Eq:GenCost_cf}
    \hat{\theta}_G = \arg \underset{\theta_G}{\min}\frac{1}{N}\sum_{k=1}^N l_{cf}(G_{\theta_G}(\mathbf{y}_k),\mathbf{x}_k)
\end{eqnarray}
In this work we will specifically design a joint loss function $l_{cf}$ as a weighted combination of several loss components that model distinct desirable characteristics of the recovered fashion image. The individual loss functions are described in more detail in the following sections.

\subsection{Generator Loss Function}
The generator loss function $l_{cf}$ can be defined as
\begin{equation}
    l_{cf} = \alpha_1 l_{MSE} + \alpha_2 l_{percept} + \alpha_3 l_{adv}
\end{equation}

\subsubsection{MSE Loss}
The mean squared error (MSE) pixel loss is defined as
\begin{equation}
    l_{MSE} = \frac{1}{WH}\|G_{\theta_G}(\mathbf{y})-\mathbf{x}\|_F^2
\end{equation}
where $W,H$ is the width and height of the images respectively, and $\|.\|_F$ is the Frobenius norm. This loss makes sure that the generated image stays close to the target image~\cite{pathakCVPR16context,XuSPZP017}. 

\subsubsection{Perceptual Loss}
Ledig et al.~\cite{Ledig2017PhotoRealisticSI} propose a perceptual loss encoder, where the ground truth image and its reconstruction are passed through a static VGG-19 network~\cite{SimonyanZ14a} and then the difference between the feature maps are obtained for different layers for the pair of images. We claim that this method is essentially computing the different filter responses learned by the network. Since these responses are obtained from a static network, we replace it by a linear set of filters encoded by the two dimensional discrete cosine transform (DCT) of the images. DCT has been used for JPEG compression for a long time and has been shown to be efficient in preserving the high frequency components of the images. The DCT based perceptual loss essentially guarantees that the reconstructed image and the original input match each other under several levels of smoothing and down-sampling operations. The simplified DCT based loss function can be written as
\begin{equation}
    l_{percept} = \frac{1}{WH}\|\Omega_{256}(G_{\theta_G}(\mathbf{y}))-\Omega_{256}(\mathbf{x})\|_F^2
\end{equation}
where $\Omega_k$ is the 2D DCT matrix of size $k \times k$. The DCT is applied channel wise to the zero padded images in our implementation.

\subsubsection{Adversarial Loss}
In addition to the content losses described so far, we also add the adversarial component of our GAN to the perceptual loss. This encourages our network to favor solutions that reside on the manifold of natural images, by trying to fool the discriminator network. The adversarial loss $l_{adv}$
is defined based on the probabilities of the discriminator $D_{\theta_D} (G_{\theta_G} (\mathbf{y}))$ over all training samples as:
\begin{equation}
    l_{adv} = \sum_{k=1}^N -\log D_{\theta_D} (G_{\theta_G}(\mathbf{y}))
\end{equation}
Here, $D_{\theta_D} (G_{\theta_G} (\mathbf{y}))$ is the probability that the reconstructed image $ G_{\theta_G} (\mathbf{y})$ is a valid reconstruction of the fashion element. For better gradient behavior we minimize $-\log D_{\theta_D} (G_{\theta_G} (\mathbf{y}))$ instead of $\log(1-D_{\theta_D} (G_{\theta_G} (\mathbf{y})))$~\cite{GAN_NIPS2014,Ledig2017PhotoRealisticSI}.

\subsection{Discriminator Loss Function}
In the recent work by Sajjadi et al.~\cite{temperedAdvNets}, the authors propose a lens network which shifts the distribution of the true images towards the generator images, at least at the beginning of the training, thereby leading to smoother training behavior for GANs. As higher epochs are reached the effect of the lens network is slowly taken off to let the network learn the true distribution of the real images. 

Based on a similar idea, we introduce a novel randomized label flipping concept which is built on the following set of equations. Let an intermediate image $I(\mathbf{x})$ be generated as
\begin{equation}\label{EQ:Lens1}
    I(\mathbf{x}) =  \beta \mathbf{x} + (1-\beta) G_{\theta_G}(\mathbf{y})
\end{equation}
where
\begin{equation}\label{EQ:beta_thresh}
    \beta \sim U(0,1) 
\end{equation}
and the label for the intermediate image $I(\mathbf{x})$ be generated as
\begin{equation}\label{Eq:LabelGen}
    \mathrm{label}~(I(\mathbf{x})) = 
  \begin{cases}
    Real~(1)      & \quad \text{if } \beta >  T\\
    Fake~(0)      & \quad \text{if } \beta \leq T
  \end{cases}
\end{equation}
where $T$ is the threshold for the process and treated as a hyper-parameter. We call this the \textbf{Randomized Label Flip} (RLF) based training.



Lower values of $T$ force the network to modify the fake images to be treated as real by the discriminator, and vice-versa.  
At lower $T$ values the generator favors all the modes present in the data due to lesser weightage being given to the fake images. As the threshold $T$ is raised, the generator starts latching onto single modes and starts matching the real data closely. The thresholding scheme along with the random sampling confuses the discriminator in two ways. The threshold measures on average how much percentage of the data combinations of the form shown in Eq.~\ref{EQ:Lens1} are labeled as fake or real. 

The randomness in sampling $\beta$ ensures that even though the decision boundary falls on one side of the threshold, there is still some randomness for the discriminator to get confused about. Our experiments show that initially starting with low values of the threshold $T$ leads to stable generator training. During the later epochs the value is raised high enough which leads to diminishing of the this behavior. 



\section{Data Generation}
Numerous fashion datasets have been published recently which have increased the enthusiasm for research into this field. DeepFashion~\cite{liuLQWTcvpr16DeepFashion} and the very recent ModaNet~\cite{zhengACMMM18modanet} are the largest fashion datasets. 
For our work, the freshness of the style trends is the most important aspect and hence we gather our own data based on web crawling for the most recent trends. We also need to make sure that the dataset can be easily updated to reflect the fashion trends of the coming seasons. We describe an end to end pipeline to generate data for learning the $c^+$GAN model proposed in the previous sections. 

\subsection{Fashion Images Scrapping}
We start by gathering large amount of images from Bing images portal\footnote{https://www.bing.com/images/}. The searches are seeded by using around 1000 fashion oriented queries like \textit{men's fall fashion}, \textit{men's fall fashion 2018}, \textit{women's fall fashion}, \textit{women's latest fall fashion} etc., similar to~\cite{liuLQWTcvpr16DeepFashion}. These leads to the first set of raw images which are fed to the next process in the pipeline. We collect around 100K images through this method.

\subsection{Fashion Model Images}
For our model to train properly, we need fashion images which show a person/model from head to toe, without missing any part. This is a crucial requirement, and we train a Faster R-CNN model~\cite{FRCNN} to detect full human body from these images. We segment all the correct detection results and pass them onto the next process in the pipleine. At this point we are left with around 80K images. 

\begin{figure*}[ht!]
\centering
\includegraphics[width=16cm,height=3.5cm]{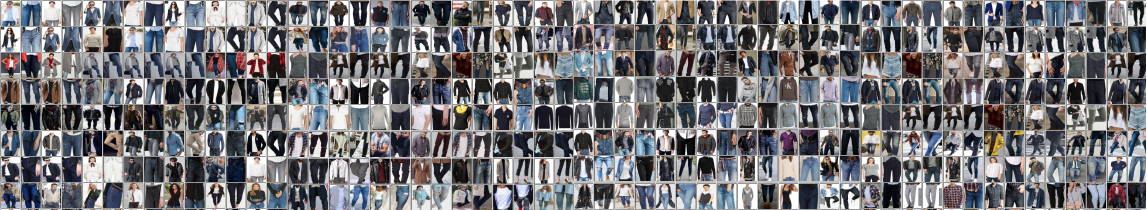}
\caption{Training pairs from our dataset. Each image from top left is paired with its bottom in the next image.}
\label{Fig:pairData}
\end{figure*}

\subsection{Segmentation to Top and Bottom Components}
\begin{figure}[h!]
\centering
\includegraphics[width=8cm]{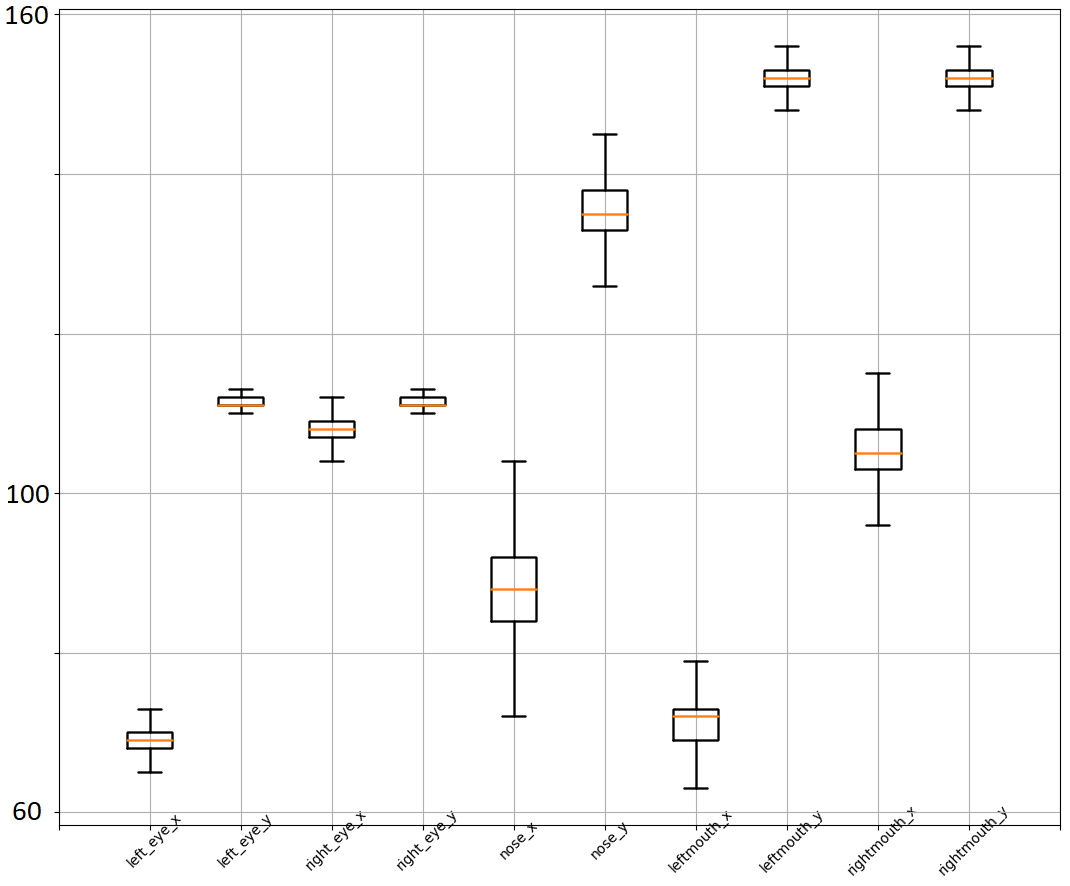}
\caption{Alignment statistics for CelebA dataset~\cite{liu2015faceattributes}. The vertical axis shows pixel locations. The first four boxes are the left and right eye co-ordinates.}
\label{Fig:StatsCelebA}
\end{figure}

\begin{figure}[ht!]
\centering
\includegraphics[width=8cm]{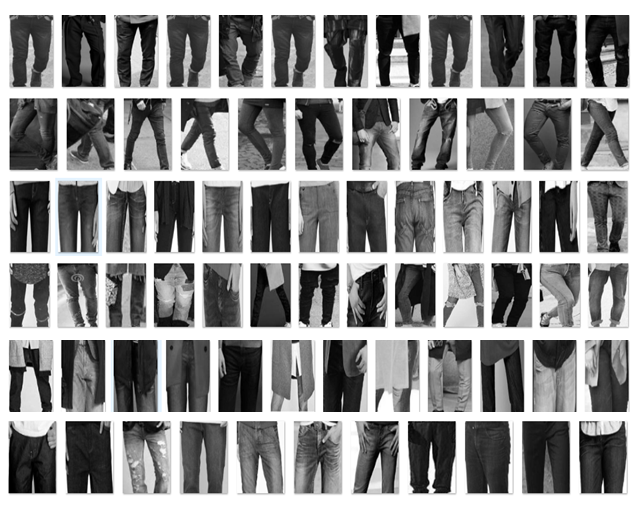}
\caption{Clustering the intensity field of the images, with K-means clustering results in these dominant clusters. Note the variation in the pose for most of the clusters. Each row is a different cluster.}
\label{Fig:DominantClusters}
\end{figure}

We provide the segmented full body images to crowd sourced workers, who segment the images into tops and corresponding bottoms. They are instructed to ignore the shoes as much as possible. The segmentation boundaries are simple rectangular axis aligned boxes for now. These components are then passed through resolution and aspect ratio checks. Note that many mistakes in the dataset still persist, such as repetitions, wrong segmentation etc. due to the errors introduced by the crowd. We let these persist through the dataset, to make the replication of this effort easier.  
We also remove non pants candidates such as skirts, frocks etc. for the results reported in this paper. Although including them is a simple addition to the data collection model, but the distribution for these is very different from the pants and trousers that we model in this paper. The methods described in this paper can be used to learn separate models for these candidates as well. These filtering operations remove more than 50\% of the data and we are left with around 35K image pairs for tops and corresponding bottoms. Some example images from our dataset are shown in Fig.~\ref{Fig:pairData}. Note that research datasets used for GANs have been heavily normalized and rectified to align them with each other. The eye co-ordinates in the CelebA dataset~\cite{liu2015faceattributes} are aligned to a couple of pixels from each other. The mean and variance for the aligned face features for the CelebA dataset are shown in Fig.~\ref{Fig:StatsCelebA}. Note that most of the GAN based face generation works report results on the aligned faces and not on the faces in the wild dateset. We perform no such alignment and the data is used as is to learn the conditional GAN model. The large amount of variability in the dataset used for our experiments is a real challenge for the current day techniques and we plan to release this data to the community to further improve the state of the art in this field. We perform a simple K-means based clustering on the intensity field of the images to get the dominant modes present in the data. Few of the dominant clusters are shown in Fig.~\ref{Fig:DominantClusters}. Note that clustering seems to identify the pose of the two legs of the pants to some degree, but does not cluster across shades. Some of the dominant clusters capture artifacts of the loosely defined labeling process such as the low hanging coat in the fifth row from the top.

\begin{figure}[ht!]
\centering
\includegraphics[width=8cm,height=3.5cm]{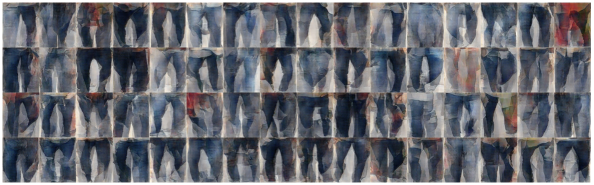}
\caption{For the basic cGAN method~\cite{MirzaO14}, the generator latches onto a combination of the dominant modes in the data during the initial epochs. This happens since the data is not aligned to begin with and hence there are no low variance regions, such as the eye corners in CelebA dataset, for the generator to latch onto during the initial epochs.}
\label{Fig:DominantGeneratorModes}
\end{figure}

The problem we are trying to address in this work is extremely hard due to the fact that the scale at which data is collected, and also the seasonality and timeliness requirements of the data make it hard to perform human annotation as well as rectification. 
Our experiments to generate the pants for given tops, while using the basic cGAN method proposed by Mirza et al.~\cite{MirzaO14} did not give expected results. Empirically, we observed that the generator for the original cGAN method latches onto the various combinations of the dominant modes present in the data as shown in Fig.~\ref{Fig:DominantGeneratorModes}, and does not converge to the ideal set of images. We hypothesize that this happens since the generator does not find any strong low variance regions, such as the eye corners for CelebA dataset, to latch onto during the initial iterations. Two possible ways to break this deadlock is either to pre-align all the training data to an average output and then generate that, or to learn a warping functional in addition to the generation cost. In this work, we take a third approach motivated by the tempered adversarial networks proposed by Sajjadi et al.~\cite{temperedAdvNets} and move the real images closer to the generated distribution to lead to more stable workload for the discriminator. We also perform a pre-clustering of the data to create semantically similar batches for the training epoch. More details of these techniques are elucidated in the following sections.

\begin{figure}[h!]
\centering
\includegraphics[width=8cm]{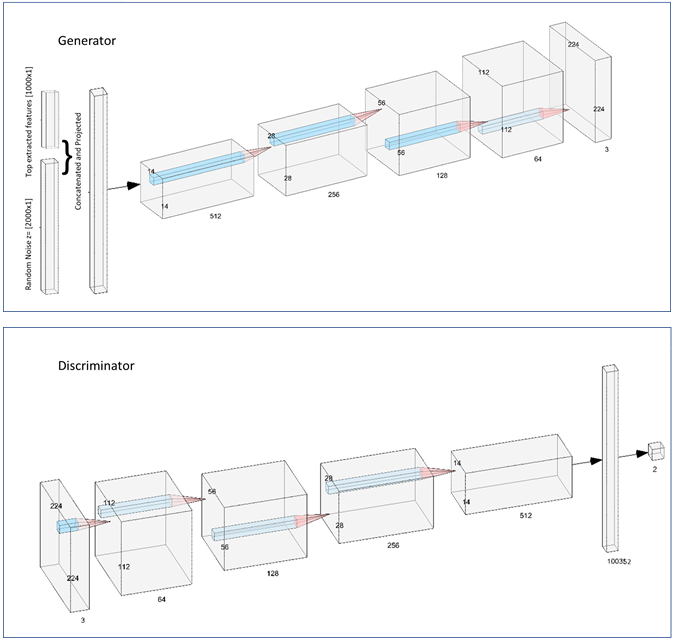}
\caption{The generator and discriminator architectures used in our work.}
\label{Fig:networkArch}
\end{figure}

\begin{figure}[h!]
\centering
\includegraphics[width=8cm,height=10cm]{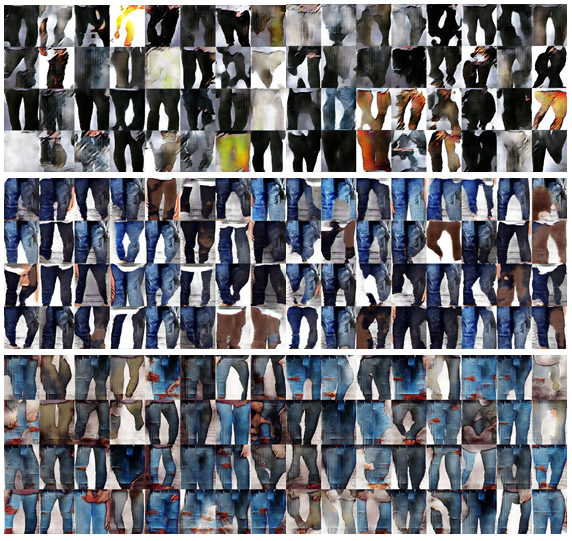}
\caption{Top panel: Adversarial only generation similar to~\cite{MirzaO14}. Middle panel: Adding MSE loss to the generator improves the output, but mixes multiple modes into individual images. Bottom panel: Adding DCT based perceptual loss brings more high frequency content, but also introduces some ringing artifacts.}
\label{Fig:withoutlens}
\end{figure}

\section{Experiments}
\subsection{Experimental Setup}
We are using conditional $c^+$GAN network for generating images. As shown in Eq.~\ref{EQ:cGANCost}, the network uses the noise vector $p_z(\mathbf{z})$ of dimension $[2000 \times 1]$ and $\mathbf{y}$ is extracted feature vector of top apparels having dimension $[1000 \times 1]$. Both these $p_z(\mathbf{z})$ and $\mathbf{y}$ are concatenated and fed to the generator network. The generator and discriminator network architectures used in our work are shown in Fig.~\ref{Fig:networkArch}.

Since we do not perform any alignment of the training data, the generator does find all the modes suitable for generation and hence keeps on generating combination of those. To counter this behavior, we converted all the input images into gray scale such that only pose variations are clustered, featurized the gray scale images and performed K-Means clustering in the feature space. This is done to separate some of the very distinct poses in the pants image dataset. 
Once this is done, the training batches are selected such that images belonging to one unique shape cluster is fed to the $c^+$GAN 
in each training batch. 


Initially the generator used only adversarial loss as proposed by Mirza et al.~\cite{MirzaO14}. 
Empirically, this does not work very well since the bottom apparels data consists of various poses, colours and designs. To accommodate all the variations, we included MSE loss and perceptual loss in generator loss function step by step. We observed that the MSE loss helped in colour matching while the perceptual loss helped in sharpening of the images. Various stages of generation has been shown in Fig.~\ref{Fig:withoutlens}. 
With only cGAN objective as proposed in~\cite{MirzaO14} the images were largely inadequate to be called apparels. This was due to the fact that the generated images had almost no texture on them. Adding MSE loss brings color matching and texture, but still high frequency details were missing. Finally, adding the perceptual loss generates the best quality images, both in variations in color and texture and also in details. 

\begin{figure}[h!]
\centering
\includegraphics[width=8cm,height=10cm]{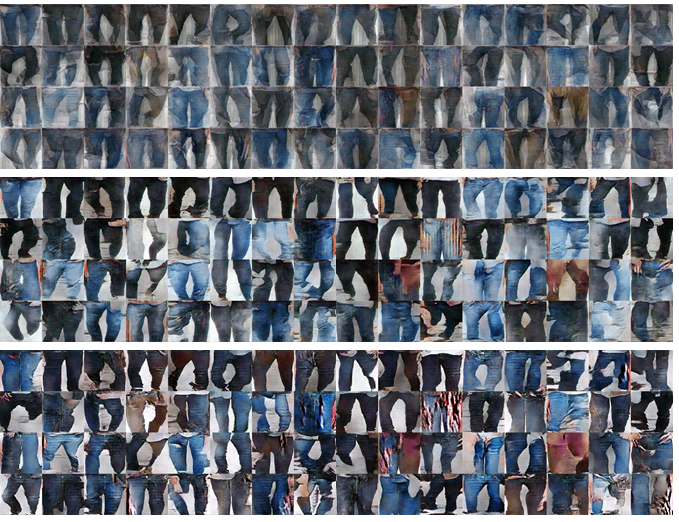}
\caption{The effect of the threshold $T$ proposed in Eq.~\ref{EQ:beta_thresh}. $T=[0.2, 0.5, 0.8]$ from top to bottom. As the threshold is increased the generator output latches onto a single mode.}
\label{Fig:multimode}
\end{figure}
\subsection{Generation using RLF training}








The generated images for lower threshold, as shown in Fig.~\ref{Fig:multimode},  provide a clear picture of how the generator is learning. Our aim for the generator is to choose the most prominent mode out of multiple modes appearing in the generated images. Once the model has trained for about 200 epochs with randomly permuted batches, we switch the training regime to clustering driven batches, where each batch contains only one dominant mode of training data. Our target pants data has lot of pose variation apart from the existing colour and design variation. At the end we need to generate pants which are of any pose but must be correct in texture and design. Switching to the clustering driven training batch regime with a high value of threshold $T=0.8$ leads to more stable generations as shown in Fig.~\ref{Fig:multimode}, bottom panel.



\begin{figure}[h!]
\centering
\includegraphics[width=8cm]{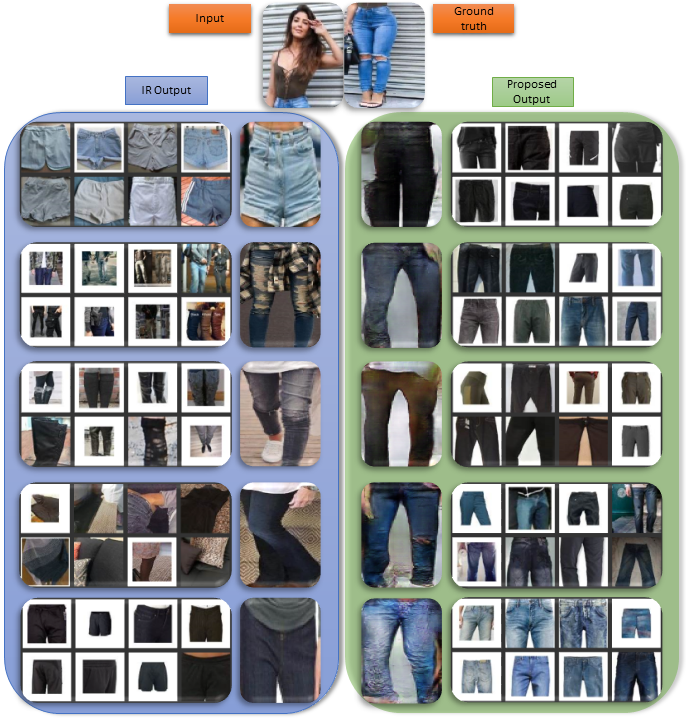}
\caption{The possible outputs for the same input as produced by the IR baseline (blue panel) and the proposed method (green panel). Note the panels on the left or right of an image are the retrievals from Bing image search for possible matches for itself. IR results are fixed and hence no improvement over this panel is possible. $c^{+}$GAN results can be sampled again and again to generate more possibilities.
}
\label{Fig:proposed5}
\end{figure}

\begin{figure*}[ht]
\centering
\includegraphics[width=16cm]{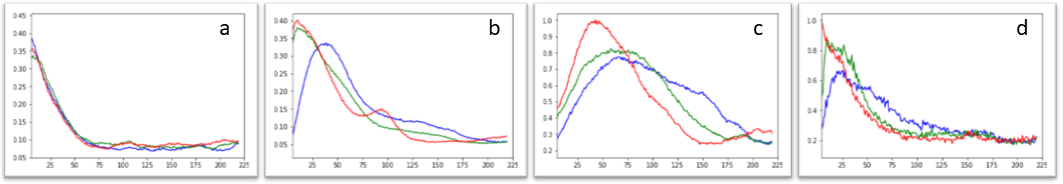}
\caption{Normalized Colour Histogram plots of batch output with a) Adversarial loss only. b) Adversarial loss + MSE loss c) Adversarial Loss + MSE loss + Perceptual loss d) RLF +  Adversarial Loss + MSE loss + Perceptual loss }
\label{Fig:colorHist}
\end{figure*}
\section{Results}

\subsection{Baseline}
We baseline out method against a simple IR baased solution, where the test clothing item, shirt in our case, is mapped to its closest shirt in the dataset and then the corresponding pants for the match are shown as the complementary recommendation. The matching algorithm used to get the closest match can be successively used to generate the top $k$ matches. 
For the results mentioned in this work, we generate Resnet50 features for the images and then generate approximate nearest neighbors by using cosine similarity over an index created using non-metric spaces library (NMSLib)\footnote{https://pypi.org/project/nmslib/}.

Note that once the matching metric and the dataset is fixed, IR based algorithm matches the same set for the given query. GAN framework, on the other hand has the amazing property to generate different results for the same input, based on the noise vector appended to the input query. This can be further exploited to correlate the noise vector with seasonality, gender, color and other such dimensions which has not been studied in this work, but remain possible extensions to our method. The comparative examples for IR based against the $c^{+}$GAN framework is shown in Fig.~\ref{Fig:proposed5}.

\subsection{Objective Evaluation}
To objectively measure the performance of our proposed method, we take inspiration from the Inception Score (IS) proposed by Salimans et al.~\cite{SalimansIS16}. IS tries to evaluate the outcome of GANs on two separate axes. 
\subsubsection{Image Quality}
The most important criteria for the generated images is whether they represent valid pants objects. We do not train a separate classifier for pants category as needed by the IS computation method. We resort to furnishing the generated images as search queries to Bing image search. If the retrieved results contain jeans / pants then we take it as good. Else we take it as a bad quality generation. We call this the search engine criteria (SEC) and report it independently. 
\subsubsection{Diversity}
\begin{figure}[ht!]
\centering
\includegraphics[width=8cm]{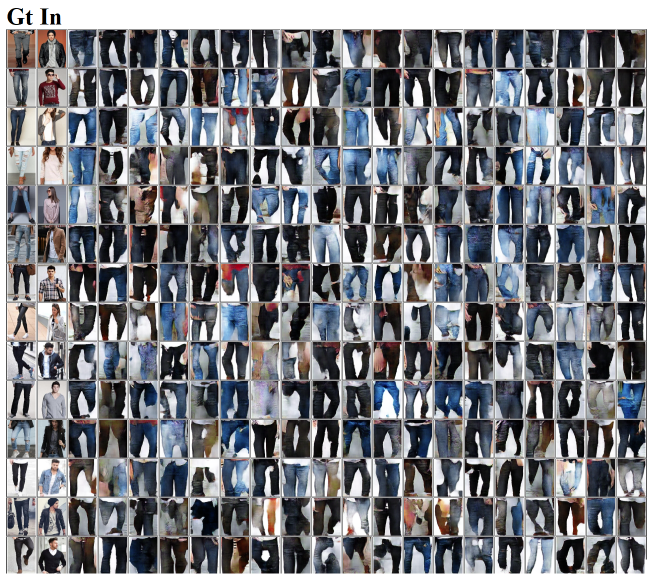}
\caption{Ground truth, Input image, and the corresponding generated outputs.}
\label{Fig:singlemode}
\end{figure}

We do not want all the images generated to look similar, which points to the generator favoring one mode over the others. We evaluate diversity by measuring the color diversity in the generated batches. To evaluate the diversity, we create a joint channel wise color histogram of all the images in a batch and compute the entropy for this distribution. A higher entropy points towards higher diversity in the generated results. In the experiments we call it the diversity criteria (DC). The histograms for each of the variants mentioned in this work is shown in Fig.~\ref{Fig:colorHist}.
The combined results for the two scores for all the variants of the proposed $c^+$GAN network are shown in Table.~\ref{table:quality}. Note that MSE loss improves SEC score, which means that more results look like jeans and pants, but addition of perceptual loss also improves the diversity in these results. Finally addition of the RLF scheme improves both these metrics over the baseline method, as well as addition of individual improvements.
Some of the images generated from our method are shown in Fig.~\ref{Fig:singlemode}. 

\begin{table}[ht] 
\centering 
\scalebox{0.8}{
\begin{tabular}{| c | c | c |} 
\hline 
\textbf{Generator Loss Method} & \textbf{SEC Score} & \textbf{DC Score}\\ [0.5ex] 
\hline 
Adversarial only  & 29.68 \% & 156.02 \\ 
Adversarial + MSE & 67.18 \% & 164.85 \\
Adversarial + MSE + Perceptual  & 64.06 \% & 193.62 \\
\textbf{RLF + Adversarial + MSE + Perceptual }      & \textbf{81.25 \%} & \textbf{212.92} \\
\hline 
\end{tabular}}
\caption{Image Quality and Diversity Score.}
\label{table:quality} 
\end{table}
\begin{figure}[ht!]
\centering
\includegraphics[width=8cm]{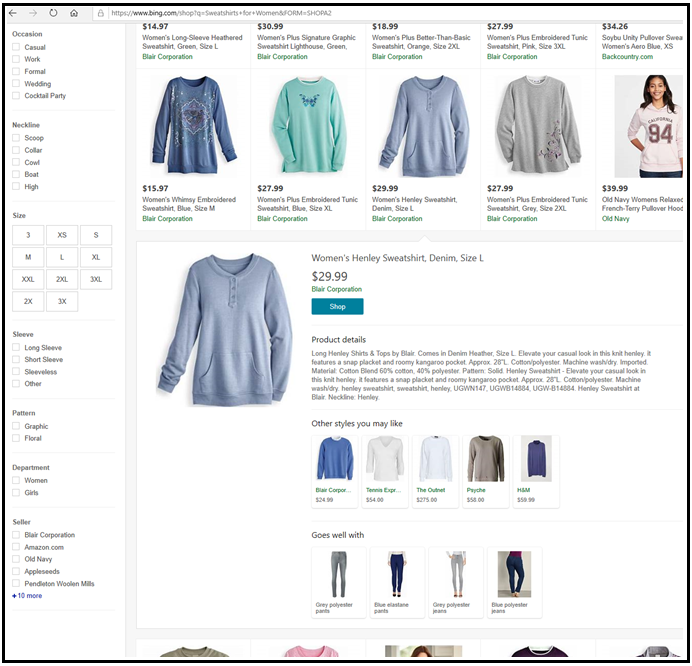}
\includegraphics[width=8cm]{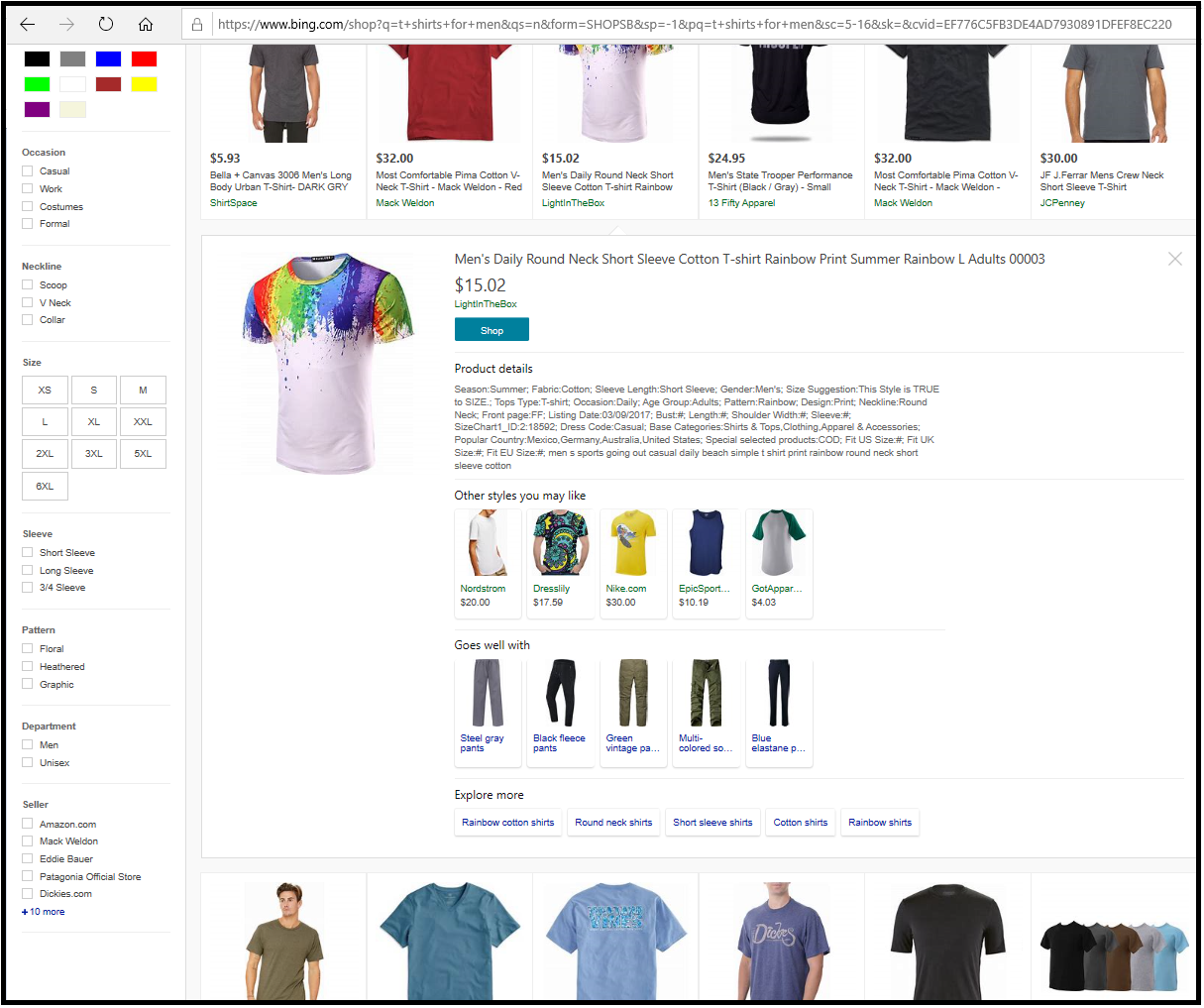}
\caption{Bing.com/shop screen grab. The feature powered by the method described in this work is shown as "Goes Well With" recommendation as the last row in the offer card.}
\label{Fig:deployed}
\end{figure}

\section{Deployment Status}
The proposed method described in this work powers the backend of the feature called `Goes Well With' in Bing Shopping vertical. The outputs generated by our method are matched against current ads corpus and then framed into a requery for the user. A screen-shot of the experience is shown in Fig.~\ref{Fig:deployed}. The live experience can be seen at www.bing.com/shop vertical for USA and UK markets, for search queries such as sweatshirts, t-shirts etc.

\section{Conclusion and Future Work}
In this work we start with a simple idea of generating relevant bottom item recommendation based on the input top item. We propose an enhanced conditional GAN model called $c^+$GAN. The proposed $c^+$GAN model modifies the generator with a classical MSE loss and also a simplified perceptual loss using DCT coefficients of the generated as well as the target images. We introduce a simplified lensing technique to the discriminator to stabilize the generator training. We cluster the data by using a very simple K-Means clustering technique which can then be used to enforce mode normalization across training batches. The final outcome of all these methods is a simple yet powerful technique which can be used to generate very meaningful fashion items which can then be used to search for similar products in e-commerce portals. Though we tackle only pants in this work, but similar techniques can be used to generate varied fashion items such as jackets, skirts etc. In the future, we would like to continue building such models to capture more categories of fashion items and also investigate the commonalities hidden in these generative models.

\bibliographystyle{ACM-Reference-Format}
\bibliography{egbib}

\end{document}